\documentclass[12pt, final]{l4dc2023}

\usepackage{caption}
\newtheorem{assumption}{Assumption}

\usepackage{soul}

\def\black#1{{\color{black}#1}}

\def\blue#1{{\color{blue}#1}}

\DeclareMathOperator*{\argmin}{argmin}

\usepackage{cleveref}
\usepackage{bm}
\usepackage{multicol}
\usepackage{mathtools}

\crefname{equation}{}{} 
\crefname{assumption}{Assumption}{}
\crefname{table}{Table}{} 
\crefname{figure}{Fig.}{}
\crefname{section}{Section}{}
\crefname{remark}{Remark}{}
\crefname{lemma}{Lemma}{}
\crefname{theorem}{Theorem}{}

\def\mcA{\mathcal{A}}

\def\mcX{\mathcal{X}}

\def\mcS{\mathcal S}

\def\mcX{\mathcal{X}}

\def \url{$u_{\text{RL}}$} 
\def \usafe{$u_{\text{safe}}$}

\def\trieq{\triangleq}

\title[Safe and Efficient RL Using Disturbance-Observer-Based CBFs]{\black{Safe and Efficient Reinforcement Learning Using Disturbance-Observer-Based Control Barrier Functions}}
\usepackage{times}

\author{%
 \Name{Yikun Cheng$^{\dagger1}$} \Email{yikun2@illinois.edu}\\
 \Name{Pan Zhao$^{\dagger1}$} \Email{panzhao2@illinois.edu}\\
 \Name{Naira Hovakimyan$^\dagger$} \Email{nhovakim@illinois.edu}\\
 \addr{$^\dagger$\textit{University of Illinois at Urbana-Champaign}}\\
 \addr{$^1$Yikun Cheng and Pan Zhao contributed equally to this work.}
}
\begin{document}
\maketitle

\begin{abstract}%

Safe reinforcement learning (RL) with assured satisfaction of hard state constraints during training has recently received a lot of attention. Safety filters, e.g., based on control barrier functions (CBFs), provide a promising way for safe RL via modifying the unsafe actions of an RL agent on the fly.  Existing safety filter-based approaches typically involve learning of uncertain dynamics and quantifying the learned model error, which leads to conservative filters  before a large amount of data is collected to learn a good model, thereby preventing efficient exploration.
This paper presents a method for safe and efficient RL using disturbance observers (DOBs) and control barrier functions (CBFs). Unlike most existing safe RL methods that deal with hard state constraints, our method does not involve model learning, and leverages DOBs to accurately estimate the pointwise value of the uncertainty, which is then incorporated into a robust CBF condition to generate
safe actions. The DOB-based CBF can be used as a safety filter with  model-free RL algorithms by minimally modifying the actions of an RL agent whenever necessary to ensure safety throughout the learning process. Simulation results on a unicycle and a 2D quadrotor demonstrate that the proposed method outperforms a state-of-the-art safe RL algorithm using CBFs and Gaussian processes-based model learning, in terms of safety violation rate, and sample and computational efficiency.
\end{abstract}

\begin{keywords}%
Reinforcement learning, robot safety, robust control, uncertainty estimation
\end{keywords}

\section{Introduction}
Reinforcement learning (RL) has demonstrated impressive performance in robotic control in recent years.  Many real-world systems are subject to safety constraints. 
As a result, safe RL has recently received a lot of attention, although there are different definitions of ``safety'' \cite{garcia2015comprehensive,brunke2022safe}. We limit our discussion to safe RL that aims to  ensure satisfaction of {\it hard state constraints} all the time
during both training and deployment.

Among different safe RL paradigms, a commonly used one is to leverage {\it safety filters} (SFs) to constrain the actions of RL agents and modify them whenever necessary to ensure satisfaction of safety constraints. The advantages of this paradigm mainly lie in its flexibility, i.e., a safety filter can often work with many existing RL algorithms without (many) modifications to the RL algorithms. Along this line, researchers have proposed different safety filters based on shielding \cite{alshiekh2018safe}, control barrier functions (CBFs) \cite{cheng2019end, ohnishi2019barrier,SafembRL2022}, Hamilton-Jacobi reachability (HJR) \cite{fisac2018general-safety}, and model predictive safety certification (MPSC) \cite{wabersich2021probabilistic-safety}. Among these different SFs, the shielding SF of \cite{alshiekh2018safe} only works for discrete state and action spaces.  
All other SFs are model-based, and hinge on Gaussian process regression (GPR) to learn the uncertain dynamics together with quantifiable prediction error for robust safety assurance. When applying these SFs to model-free RL,  model learning is still needed. More importantly, due to the reliance on model learning, when the learned model is poor due to insufficient data, existing model-based SFs will be overly conservative, preventing efficient exploration of RL agents. Additionally, it is well known that GPR is computationally demanding (standard GPR model training involves computing the inverse of an $N\times N$ covariance matrix, where $N$ is the number of data points). As a result, GPR is probably not scalable for disturbance estimation in high-dimensional systems, particularly compared to alternative methods such as disturbance observers. 

\begin{figure}
\vspace{-12mm}
    \centering
    \includegraphics[width=0.6\linewidth]{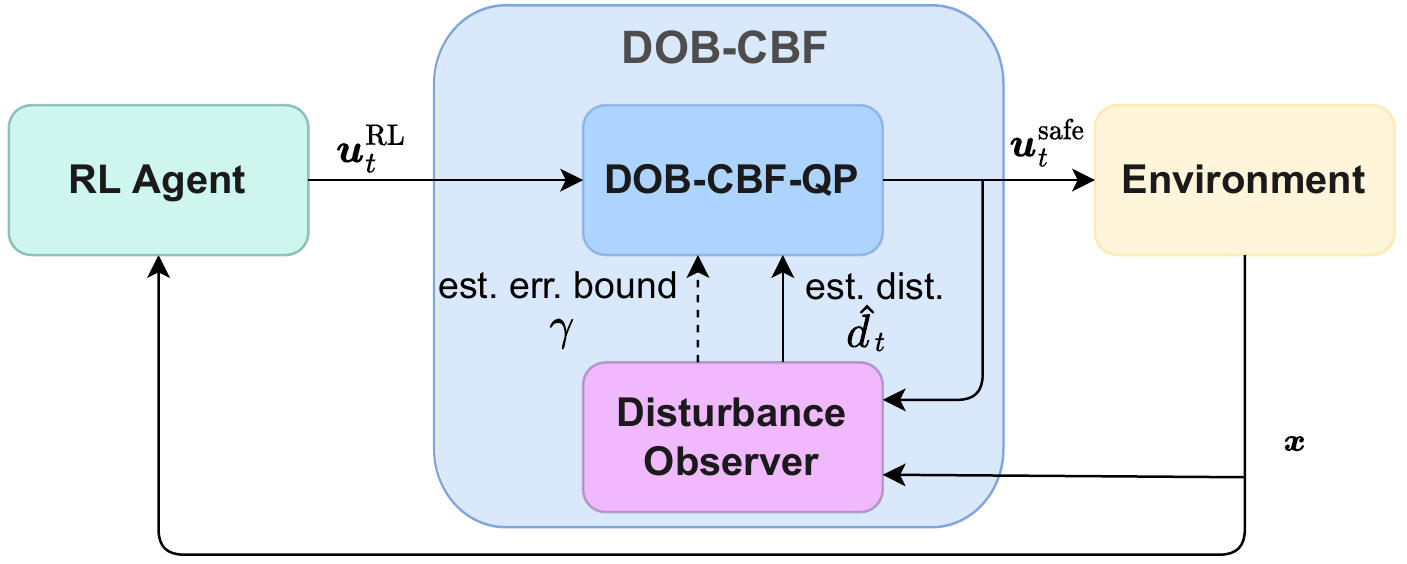}
    \caption{Proposed safe RL framework via using DOB-CBF. At each time step $t$, the RL agent action $u^{\text{RL}}_t$ potentially violates the predefined safety constraints. A DOB-CBF based safety filter will render a safe action $u^{\text{safe}}_t$ based on $u^{\text{RL}}_t$, a precomputed estimation error bound $\gamma$ and  an estimated disturbance $\hat{d}_t$. Then, $u^{\text{safe}}_t$ is applied to interact with the environment to enforce safety during and after the policy training.}
    \label{fig:framework}
    \vspace{-5mm}
\end{figure}
This paper presents a safe RL approach using disturbance observer (DOB) based robust CBFs that were  introduced in \cite{zhao2020aR-cbf}, and later extended in \cite{dacs2022robust-dob-cbf}. As illustrated in Figure \ref{fig:framework},  our approach leverages a DOB to accurately estimate the value of the lumped disturbance at each time step with a pre-computable estimation error bound (EER). The estimated disturbance together with the EEB is incorporated into a quadratic programming (QP) module with robust CBF conditions that generates safe actions at each step by minimally modifying the RL actions. 
Compared to existing CBF-based safe RL approaches, e.g., \cite{cheng2019end, ohnishi2019barrier,SafembRL2022}, our approach does not need model learning of the uncertain dynamics (although a nominal model is needed). 

This article is organized as follows. Section \ref{preliminary} includes preliminaries related to  DOBs, CBFs,  RL. Section \ref{main_approach} presents the proposed safe RL framework, while Section \ref{simulation} includes the simulation results for verifying the proposed framework using a unicycle and a 2D quadrotor.  


\section{Preliminaries} \label{preliminary}
We consider a nonlinear control-affine system 
\begin{equation}{\label{eq:dynamics}}
    \dot{x}(t) = f(x(t)) + g(x(t))u(t) + d(x(t)),
\end{equation}
where $x(t) \in \mathcal{X} \subset \mathbb{R}^n$ denotes the state vector, $u(t) \in \mathcal{U} \subset \mathbb{R}^m$ is the input vector, $\mathcal{X}$ and $\mathcal{U}$ are compact sets, $ f: \mathbb{R}^n \to \mathbb{R}^n$ and $g: \mathbb{R}^n \to \mathbb{R}^m$ are known and locally Lipschitz-continuous functions, and $d: \mathbb{R}^n \to \mathbb{R}^n$ is an unknown function that captures uncertain dynamics.
\begin{assumption}\label{assump:lipschitz-bound-d}
There exist positive constants $l_d$ and $b_d$ such that for any $x,y\in \mathcal{X}$, the following inequalities hold:
\begin{align}
\|d(x)-d(y)\| & \leq l_d\|x-y\|, \label{eq:Lipschitz-bnd}\\
\|d(0)\| & \leq b_d.\label{eq:uniform-bnd}
\end{align}
\end{assumption}
Moreover, the constants $l_d$ and $b_d$ are known. 

\begin{remark} 
 \cref{assump:lipschitz-bound-d} is a fairly standard assumption in nonlinear systems stating that the uncertain function $d(x)$ is {\it locally Lipschitz} continuous with a known {\it bound}  on the Lipschitz constant {\it in the compact set $\mcX$} and is bounded by a known constant at the origin. \black{The Lipschitz constant and uniform bounds will affect the estimation error bound when using a DOB to estimate the uncertainty $d(x)$, as characterized in Lemma~\ref{lemma:est_error_bound}}. 
\end{remark}

\subsection{Reinforcement Learning}
Reinforcement learning aims to find an optimal policy $\pi^{*}$ in an environment which can be formulated as a Markov decision process (MDP). In this work, an MDP is defined by a tuple $(\mathcal{S}, \mathcal{A}, p, r)$, where the state space $\mathcal{S}$ and the action space $\mathcal{A}$ are continuous. Given the current state $s_{t}\in \mcS$ and action $a_{t}\in\mcA$,  the transition function $p: \mathcal{S} \times \mathcal{S} \times \mathcal{A} \to [0,\infty)$ represents the probability density of the succeeding state $s_{t+1}\in\mcS$. The reward function $r: \mcS\times\mcA\to [r_\textup{min},r_\textup{max}]$ determines a bounded reward for each transition. 

Our proposed safe RL scheme can work with any model-free RL algorithm. For illustration and experimental demonstration in \cref{simulation}, we choose soft actor-critic (SAC) \cite{Tuomassac},  a state-of-the-art model-free RL algorithm. 
SAC uses an off-policy formulation that reuses historical data to improve sample efficiency and utilizes entropy maximization to improve the stability of the training process. In general, SAC aims to find a policy to maximize an entropy objective which is formed as $\sum_{t=0}^{T} \mathbb{E}_{\left(x_t, a_t\right) \sim \rho_\pi}\left[r\left(x_t, a_t\right)+\alpha \mathcal{H}\left(\pi\left(\cdot \mid x_t\right)\right)\right]$, where $\mathcal{H}(\cdot)$ is the entropy term that incentivizes exploration, $\alpha$ is a positive parameter to determine the relative importance of the entropy term against the reward, $\rho_\pi$ denotes the states and actions distribution induced by the policy $\pi$, and $T$ is the termination time.
\subsection{Control Barrier Function}
The CBFs are introduced in \cite{ames2016cbf-tac} to synthesize control laws to ensure forward invariance of some sets (often related to safety) for nonlinear control-affine systems. They are often used as safety filters to modify a baseline control law to ensure that the system stays in a safety set. 
Consider a set 
\begin{equation}\label{eq:forward_invariant}
    \mathcal{C} := \{x \in \mathbb{R}^n : h(x)\geq 0\} \subseteq \mcX,
\end{equation}
where $h(x)$ is a continuously differentiable function $h$. A function $\beta:(-b, a)\to (-\infty, \infty)$ is said to belong to extended class $\mathcal{K}$ for some $a,b > 0$ if it is strictly increasing and $\beta(0)=0$.
\begin{definition}{\label{def:rcbf}}(CBF \cite{ames2016cbf-tac}).
Given a set $\mathcal{C}$ defined using $h(x)$ via (\ref{eq:forward_invariant}), $h(x)$ is a control barrier function for (\ref{eq:dynamics}) if there exists an extended class $\mathcal{K}$ function $\beta$ such that $\forall x \in \mathcal{C}$
\begin{equation}\label{eq:cbf}
    \sup _{u \in \mathcal{U}}\left\{L_f h(x)+L_g h(x) u + h_{x}(x)d(x)\right\} \geq-\beta(h(x)),
\end{equation}
\end{definition}
where $h_x(x)\triangleq\frac{\partial h(x)}{\partial x}$, $L_fh(x)\triangleq \frac{\partial h(x)}{\partial x}f(x)$ and $L_gh(x)\triangleq \frac{\partial h(x)}{\partial x}g(x)$. 

We define the input relative degree (disturbance relative degree) of a differentiable function $h: \mathbb{R}^n \to \mathbb{R}$ with respect to (\ref{eq:dynamics}) as the number of times we need to differentiate it along  (\ref{eq:dynamics}) until the input $u$ (the disturbance $d$) explicitly shows up. Condition \cref{eq:cbf} works only for constraints with input relative degrees (IRDs) of one. To handle constraints with higher IRDs, high-order CBFs are introduced in \cite{xiao2021high-order-cbf}. Before introducing high-order CBFs, we make the following assumption, which indicates that the input $u$ and the disturbance $d$ show up together when differentiating $h$. 
\begin{assumption}\label{assump:disturb-rel-degree}
The disturbance relative degree is equal to the input relative degree.
\end{assumption}
Define a sequence of functions $\phi_{i}: \mathbb{R}^n \to \mathbb{R}, i \in \{1,...,m\}$ as:
\begin{equation}{\label{eq:phi}}
    \phi_{i}(x) = \dot{\phi}_{i-1}(x) + \beta_{i}(\phi_{i-1}(x)), \phi_{0} = h(x).
\end{equation}
Furthermore, define an associate sequence of sets as:
\begin{equation}{\label{eq:sets-c}}
    \mathcal{C}_i=\left\{x \in \mathbb{R}^n: \phi_{i-1}(x) \geq 0\right\} \subseteq \mcX,~ i \in\{1, \ldots, m\}.
    \vspace{-9mm}
\end{equation}
\begin{definition}{\label{def-horcbf}} (High-Order CBF under Perturbed System Dynamics).
Consider a sequential function $\phi_{i}(x)$ defined in (\ref{eq:phi}) and a sequential set $\mathcal{C}_i, i \in \{1,...,m\}$ defined in (\ref{eq:sets-c}). Under \cref{assump:disturb-rel-degree}, an $m^{th}$-order differentiable function $h: \mathbb{R}^n \to \mathbb{R}$ is a high-order CBF of IRD m for (\ref{eq:dynamics}) if there exist extended differentiable class $\mathcal{K}$ functions $\beta_{i}, i \in \{1,...,m\}$, such that $\forall x \in \mathcal{C}_{1}\cap ..., \cap \mathcal{C}_{m}$
\begin{equation}
    \sup _{u \in \mathcal{U}}\mathcal{L}_f^m h(x)+\mathcal{L}_g \mathcal{L}_f^{m-1} h(x) u+[\mathcal{L}_f^{m-1} h(x)]_{x} d(x)+O(h(x))+\beta_m\left(\phi_{m-1}(x)\right) \geq 0\label{eq:horcbf}, 
\end{equation}
where $L^{m}_fh(x)=\frac{\partial L_f^{m-1} h(x)}{\partial x} f(x)$, $\mathcal{L}_g \mathcal{L}_f^{m-1} h(x)=\frac{\partial L_f^{m-1} h(x)}{\partial x} g(x)$ and $[\mathcal{L}_f^{m-1} h(x)]_{x} = \frac{\partial L_f^{m-1} h(x)}{\partial x}$, and  $O(h(x))=\sum_{i=1}^{m-1} \mathcal{L}_f^i\left(\beta_{m-i} \circ \phi_{m-i-1}\right)(x)$.
\end{definition}
The true uncertainty $d$ in Definitions \ref{def:rcbf} and \ref{def-horcbf} is not accessible in practice. Therefore, it is impossible to evaluate whether a function $h(x)$ obeys the constraints in (\ref{eq:cbf}) or (\ref{eq:horcbf}).
One solution is to derive a sufficient condition for (\ref{eq:cbf}) or (\ref{eq:horcbf}) using a uniform bound for the uncertainty $d(x)$, as adopted in \cite{zhao2020aR-cbf,nguyen2016optimalACC}. In the following, we will derive an alternative sufficient condition  to define the so-called DOB-CBFs. 
\subsection{Disturbance Observer (DOB) with a Precomputable Estimation Error Bound}
Disturbance observers have been widely used in control of uncertain systems
\cite{chen2015dobc}. All different types of DOBs share a common idea, i.e., lumping all the uncertainties (that may consist of unknown parameters, unmodeled dynamics and external disturbances) together as a total disturbance and estimate its value at each time instant. In this work,  we leverage the DOB presented in \cite{zhao2020aR-cbf}, which is inspired by the piecewise-constant (PC) adaptive law \cite{wang2017adaptiveMPC} and \cite[Section 3.3]{naira2010l1book-nh}, and was used in $\mathcal L_1$ adaptive control of manned aircraft \cite{ackerman2017evaluation,ackerman2019Learjet}, and in learning-enabled control \cite{gahlawat2020l1gp,cheng2022improvingRL-ral}. 
The DOB contains two components, i.e., a state predictor and a PC estimation law to estimate the disturbance. For the disturbed system (\ref{eq:dynamics}), the state predictor is given by
\begin{equation}\label{eq:state-predictor}
    \dot{\hat{x}}(t) = f(x) + g(x)u + \hat{d}(t) - a\Tilde{x},
\end{equation}
where $\Tilde{x} = \hat{x} - x$ denotes the prediction error, $a > 0$ is a constant, and $\hat{d}(t)$ is the estimated disturbance. The disturbance estimation is updated according to  
\begin{equation}{\label{eq:pc-law}}
    \left\{\begin{aligned}
    \hat{d}(t) &=\hat{d}(i T), \quad t \in[i T,(i+1) T), \\
    \hat{d}(i T) &=-\frac{a}{e^{a T}-1} \tilde{x}(i T), i=0,1,...,
    \end{aligned}\right.
\end{equation}
where $T$ is the estimation sampling time. 
The estimation error bound associated with the DOB is given next.
\begin{lemma} (Estimation Error Bound \cite{zhao2020aR-cbf})\label{lemma:est_error_bound}
Given the uncertain system (\ref{eq:dynamics}) subject to Assumption \ref{assump:lipschitz-bound-d}, and the DOB defined via (\ref{eq:state-predictor}) and (\ref{eq:pc-law}), the estimation error can be bounded as
\begin{equation}
\begin{aligned}
\|\hat{d}(t)-d(x(t))\| &\leq{\delta(t)} \trieq \begin{cases}\theta \triangleq l_d \max _{x \in \mathcal{X}}\|x\|+b_d, & \forall 0 \leq t<T, \\ \gamma(T) \triangleq 2 \sqrt{n} \eta T+\sqrt{n}\left(1-e^{-a T}\right) \theta, & \forall t \geq T,\end{cases}\label{eq:error-bound}
\end{aligned}
\end{equation}
\end{lemma}
where $\eta\triangleq l_d(\max _{x \in \mathcal{X}, u \in \mathcal{U}}\|f(x)+g(x) u\|+\theta)$. Moreover, $\lim _{T \rightarrow 0} \gamma(T)=0$.
\begin{remark}
Lemma~\ref{lemma:est_error_bound} implies that the estimated disturbance can be made arbitrarily accurate for $t\geq T$, by reducing $T$, the latter only subject to hardware limitations, including measurement noises. 
\vspace{-3mm}
\end{remark}

\black{\begin{remark}
As seen from Lemma~\ref{lemma:est_error_bound}, larger values of the Lipschitz bound $l_d$ and the uniform bound $b_d$ (introduced in \cref{assump:lipschitz-bound-d}) lead to larger estimation error bounds. 
On the other hand, we found that the estimation error bound in \cref{eq:error-bound} could be quite conservative in most cases, potentially due to the frequent use of Lipschitz continuity properties and triangular inequalities. Therefore, for practical implementation one
could leverage some empirical study, e.g., doing simulations
under a few user-selected functions of $d(x)$ and determining a tighter bound than that defined in \cref{eq:error-bound}. The conservativeness also indicates that even if the constants $l_d$ and  $b_d$ violate the conditions \cref{eq:uniform-bnd} to a certain extent, the error bound  in \cref{eq:error-bound} most likely still holds in practice. 
\vspace{-3mm}
\end{remark}}

\section{Main Approach}\label{main_approach}
In this section, we first introduce high-order DOB-based CBFs (DOB-CBFs), by extending the result in \cite{zhao2020aR-cbf}, which only considers constraints with IRDs of one. The work in \cite{dacs2022robust-dob-cbf} considers high IRD constraints using exponential CBFs in the presence of matched uncertainties (which are injected to the system through the same channel as control inputs). Compared to \cite{dacs2022robust-dob-cbf}, we do not constrain the uncertainties to be matched, and leverage high-order CBFs \cite{xiao2021high-order-cbf} which are generalizations of exponential CBFs.
Then, we introduce our DOB-CBF based safe RL scheme (DOB-CBF-RL).
\subsection{High-Order DOB-Based Control Barrier Function (DOB-CBF)} \label{section-pc}
Given the disturbance bound in \cref{assump:lipschitz-bound-d} and the estimation error bound in Lemma \ref{lemma:est_error_bound}, 
we have  $[\mathcal{L}_f^{m-1} h(x)]_{x}d(x)\leq\|[\mathcal{L}_f^{m-1} h(x)]_{x}\|(\theta +2\delta(t))$. The detailed proof could be found in the extended version of this paper \cite{cheng2022safe}.
Therefore, we have the following definition.

\begin{definition} (DOB-CBF)\label{defn:dob-cbf}
\black{Consider the system in \cref{eq:dynamics},  the DOB defined via \cref{eq:state-predictor,eq:pc-law} with an estimation error bound given by \cref{eq:error-bound}, and a sequential function $\phi_{i}(x)$ defined in (\ref{eq:phi}) and a sequential set $\mathcal{C}_i, i \in \{1,...,m\}$ defined in (\ref{eq:sets-c}).} Under Assumption \ref{assump:disturb-rel-degree}, an $m^{th}$-order differentiable function $h: \mathbb{R}^n \to \mathbb{R}$ is a high-order DOB-based control barrier function of relative degree m for (\ref{eq:dynamics}) if there exist extended class $\mathcal{K}$ functions $\beta_{i}, i \in \{1,...,m\}$, such that
 \vspace{-2mm}
 \begin{equation}\label{worst-bound}
     \sup _{u \in \mathcal{U}}\mathcal{L}_f^m h(x)+\mathcal{L}_g \mathcal{L}_f^{m-1} h(x) u-\|[\mathcal{L}_f^{m-1} h(x)]_{x}\| (\theta+2\max_{t\in[0,\infty)}\black{\delta(t)})+O(h(x))+\beta_m\left(\phi_{m-1}(x)\right) \geq 0,
 \end{equation}
 for all $x \in \mathcal{C}_{1}\cap ..., \cap \mathcal{C}_{m}$.
\end{definition}
It is obvious that if a control input $u$ is a solution for (\ref{worst-bound}), it also satisfies (\ref{eq:horcbf}). We next define

\vspace{-5mm}
\begin{equation}\label{cond1}
    \mathcal{K}(t, x, u) \triangleq L^{m}_f h(x)+L_gL^{m-1}_f h(x) u+O(h(x))+[\mathcal{L}_f^{m-1} h(x)]_{x} \hat{d}(t)-\|[\mathcal{L}_f^{m-1} h(x)]_{x}\| \black{\delta(t)}.
\end{equation}
Then, the main theorem of the proposed approach is introduced as follows.
\begin{theorem}{\label{theorem1}}
Suppose the condition (\ref{worst-bound}) holds. Then, the condition
\vspace{-1mm}
    \begin{equation}
        \sup _{u \in \mathcal{U}} \mathcal{K}(t, x, u) \geq-\beta_m\left(\phi_{m-1}(x)\right){\label{condition-thm1}}
    \end{equation}
is a sufficient condition for (\ref{eq:horcbf}). It is also a necessary condition for (\ref{eq:horcbf}) for any $t\geq T$ when $T \to 0$.
\end{theorem} 
The proof of Theorem \ref{theorem1} could be found in \cite{cheng2022safe}.

\subsection{Safe RL Policy Training with DOB-CBFs}
Using the condition in (\ref{condition-thm1}) that depends on the estimated disturbance $\hat{d}(t)$, we can compute the safe control inputs via solving a quadratic programming (QP) problem defined as
\vspace{-3mm}
\begin{equation}
    \begin{aligned}{\label{eq:DOB-CBF-QP}}
    u_{\text{safe}}&=\argmin _{u\in \mathbb{R}^{m}} \frac{1}{2}(u-u_{\text{RL}})^{T}P(u-u_{\text{RL}}) ~(\textbf{DOB-CBF-QP})\\
    \text{s.t.} & ~\mathcal{K}(t, x, u) +\beta_m\left(\phi_{m-1}(x)\right) \geq 0, \\
    & ~u \in \mathcal{U},
    \end{aligned}
\end{equation}
where $P$ is a positive-definite weighting matrix,   $u_{\text{RL}}$ is the action of RL policy and $u_{\text{safe}}$ is the final control input applied to the system (\ref{eq:dynamics}) during both policy training and policy deployment. In case \url satisfies the constraints of \cref{eq:DOB-CBF-QP} and is therefore safe, we have \usafe$=$\url; otherwise, \cref{eq:DOB-CBF-QP} produces safe inputs that are mostly close to \url.
\begin{algorithm2e}
\caption{DOB-CBF based safe RL}\label{alg:1}
\SetKwInOut{Input}{Input}
\Input{Initial \blue{SAC} policy $\pi_\theta$, number of episodes $N$, number of steps per episode $M$, number of policy update $G$,  nominal dynamics $\dot{x} = f(x)+g(x)u$, DOB defined via \cref{eq:state-predictor} and \cref{eq:pc-law} with estimation error bound $\gamma$,}
\For{$i=1,....,N$}{
    \For{$t=1,...,M$}{
        Obtain action $u^{\text{RL}}_t$ from policy $\pi_{\theta}$\\
        Obtain disturbance estimation $\hat{d}_t$  from the DOB defined via \cref{eq:state-predictor} and \cref{eq:pc-law}\\
        Obtain safe action $u^{\text{safe}}_t$ from DOB-CBF-QP defined in \cref{eq:DOB-CBF-QP}, using $u^{\text{RL}}_t$, $\gamma$ and $\hat{d}_t$\\
        Take action $u^{\text{safe}}_t$ in the environment\\
        Add transition $(x_t, u^{\text{safe}}_t, x_{t+1}, r_t)$ to replay buffer $\mathcal{D}$\\
        \For{$j=1,...,G$}{
        Sample mini-batch $\mathcal{B}$ from $\mathcal{D}$\\
        Update policy $\pi_\theta$ using $\mathcal{B}$}
    }
}
\end{algorithm2e}
\setlength{\textfloatsep}{0pt}
With the DOB-CBF-QP in (\ref{eq:DOB-CBF-QP}), our proposed DOB-CBF-RL scheme is summarized in Algorithm \ref{alg:1}.
At each step during training, the vanilla RL policy determines a potentially unsafe action. This action is then modified by the DOB-CBF-QP in \cref{eq:DOB-CBF-QP} to produce a safe action $u_{\text{safe}}$ that is applied to the environment. It is worth mentioning that the tuple $(x_t, u^{\text{safe}}_t, x_{t+1}, r_t)$ involving the safe action is added to the reply buffer $\mathcal{D}$ and used to update the policy. Using safe actions for policy training will promote the agent to learn a safe and optimal policy (although not guaranteed), which the DOB-CBF as a safety filter does not need to (frequently) intervene with. 

\section{Simulation}\label{simulation}
We use a unicycle and a 2D quadrotor to validate the efficacy of the proposed DOB-CBF-RL method. For comparison,  we also implemented  a state-of-the-art safe RL method based on CBFs and GPR-based model learning (denoted as GP-CBF-RL) in \cite{cheng2019end}. The policy training was performed on a desktop with an INTEL i9-9980XE and an NVIDIA 3090Ti GPU and 64GB RAM. \black{\black{The codes are available at \hyperlink{https://github.com/Adriancyk/Safe-RL-with-Disturbance-Observer}{https://github.com/Adriancyk/Safe-RL-with-Disturbance-Observer}.}}

\subsection{Unicycle}
\begin{figure}[h]
\vspace{-2mm}
    \centering
    \includegraphics[scale=0.39]{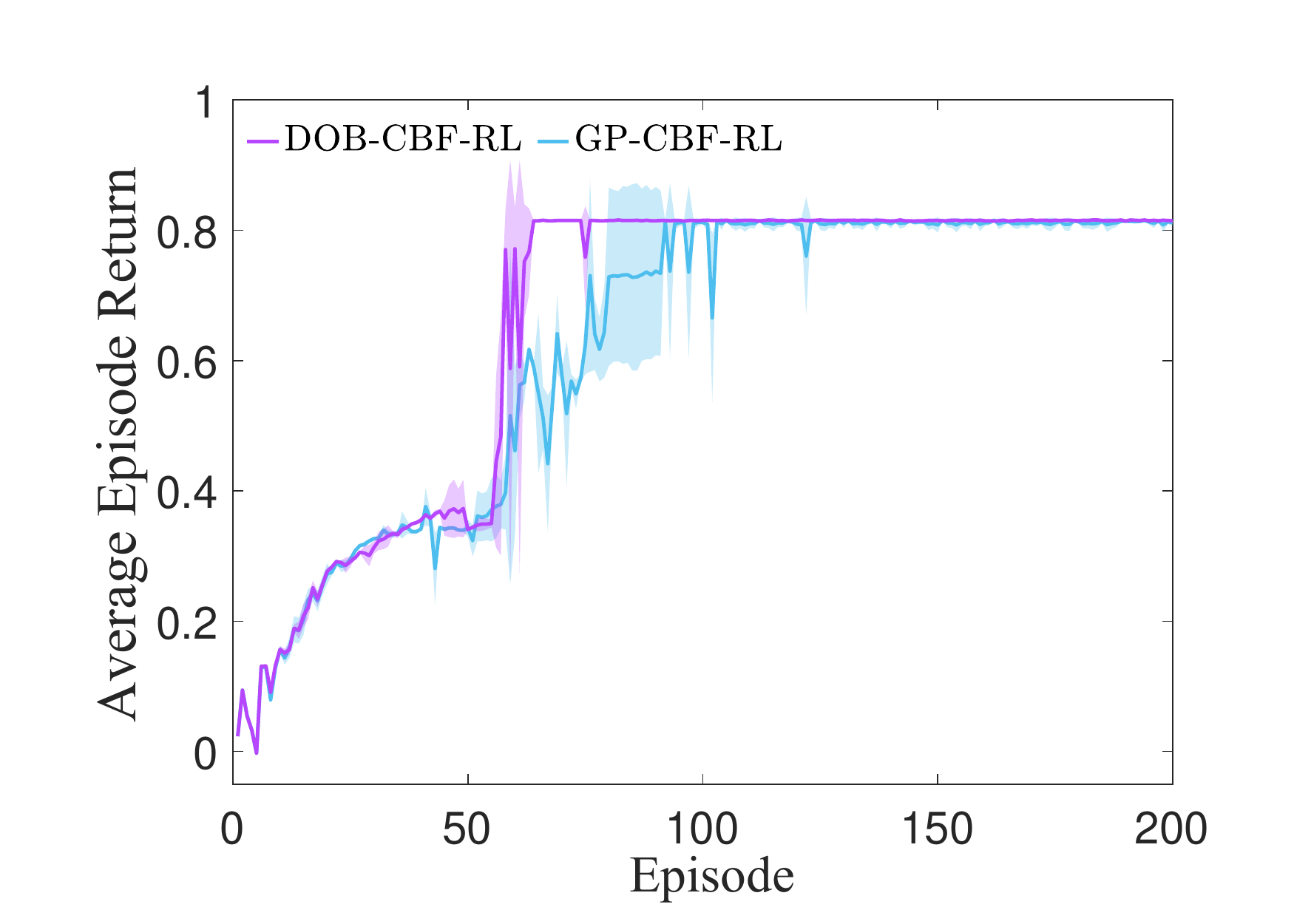}
    \includegraphics[scale=0.39]{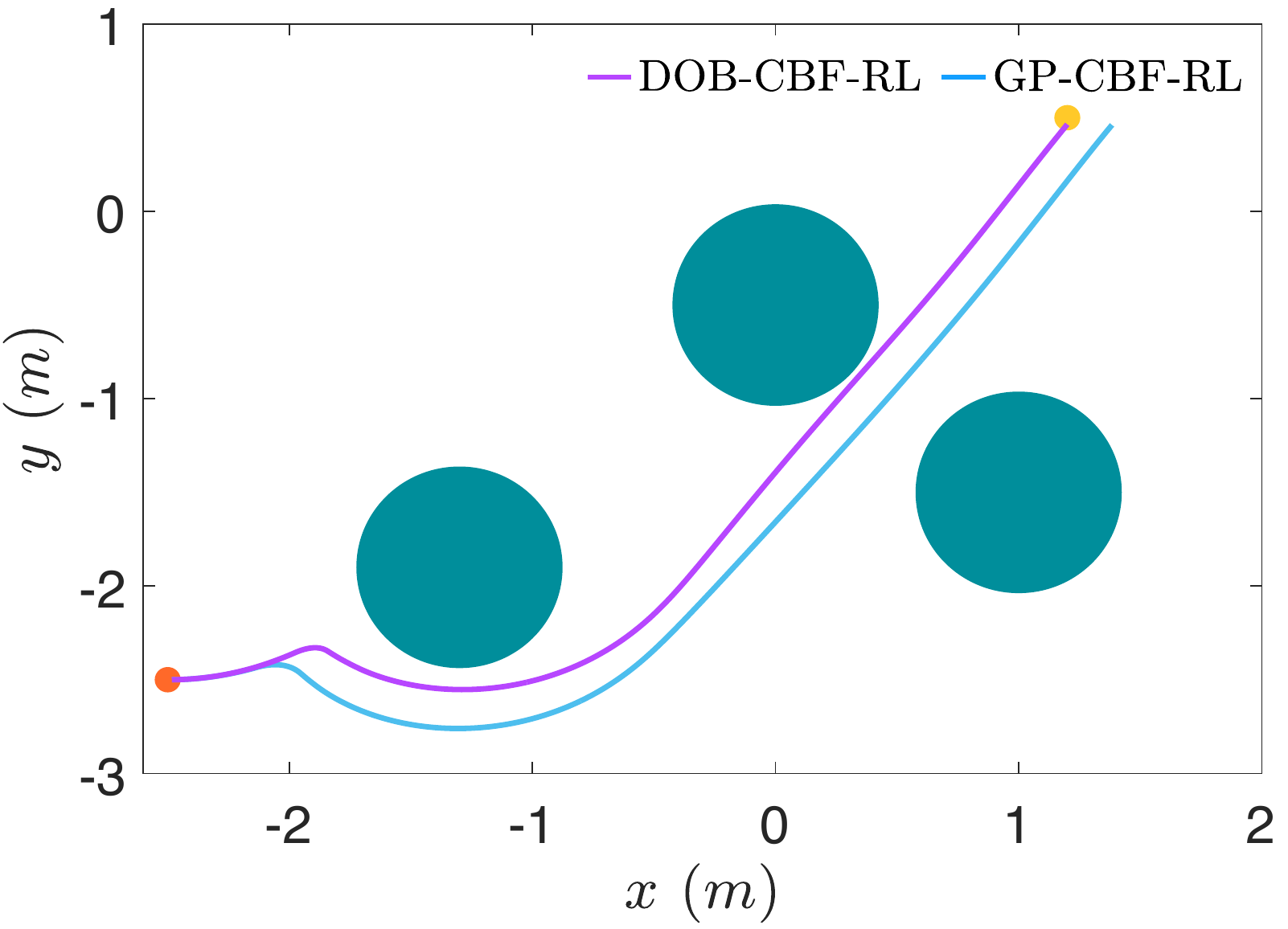}
    \vspace{-2mm}
\caption{(Left) Unicycle training curves for DOB-CBF-RL and GP-CBF-RL. The solid lines and shaded areas denote the mean and standard deviation over five trials. (Right) Navigation performance for DOB-CBF-RL policy and GP-CBF-RL policy trained in 200 episodes.}\label{fig:uni_train_traj}
\end{figure}

A unicycle model was borrowed from \cite{SafembRL2022} and adapted. The state $x = [p_x, p_y, \theta]^T$, where $p_x$ and $p_y$ denote the robot position along the x-axis and y-axis, respectively, and  $\theta$ is the counterclockwise angle between the positive direction of the x-axis and the head direction of the robot. The control inputs are the linear velocity $v$ and angular velocity $\omega$ of the system. The goal is to navigate the unicycle from the red dot to the yellow dot without colliding with any obstacles, as shown in Figure \ref{fig:uni_train_traj} (Right). The vehicle's equations of motion are given by
\begin{equation}{\label{eq:uni_dyn}}
    \begin{aligned}
    v_x &=\cos{\theta}(v + d_{m}), \\
    v_y &=\sin\theta(v + d_{m}), \\
    \dot{\theta} &=\omega,
    \end{aligned}
\end{equation}
\black{where the uncertainty $d_m$ is introduced to mimic the slippery ground that causes the vehicle to lose partial control efficiency}. 
We train DOB-CBF-RL policy and the GP-CBF-RL policy separately and define $h_i(x) = \frac{1}{2}((\|p_{i,\textup{obs}}\|-\sqrt{p^2_x+p^2_y})^2-r_{i,\textup{obs}}^{2})$, where $i=1,2,3$, $p_{i,\textup{obs}}$ denotes the location of $i$th obstacle in xy-plane, and $r_{i,\textup{obs}}$ is the radius of the $i$th obstacle. 
The parameters of the system dynamics and the details of the RL training are included in \cite{cheng2022safe}. 

We can see from Figure \ref{fig:uni_train_traj} (Left) that the DOB-CBF-RL policy consistently converges within 60 episodes. In comparison, it takes at least 90 episodes for the GP-CBF-RL to find an equally good policy. Considering the use of 200 episodes to train a policy, a 30-episode gap is a considerable improvement in training efficiency. It is worth noting that compared with DOB-CBF-RL, there are more variations in the training performance across different trials under GP-CBF-RL. It indicates that DOB-CBF-RL can further improve the training stability by providing accurate disturbance estimation. Figure \ref{fig:uni_train_traj} (Right) compares the navigation performance. DOB-CBF-RL provides a more aggressive way to approach the target. In comparison, GP-CBF-RL chooses a more conservative trajectory and fails to reach the target, although the deviation is negligible. The safety violation rate for each 50 training episodes during training is listed in Table \ref{table:safety_violation_uni}. One can see that DOB-CBF-RL achieves zero-violation rates during the entire training process, while GP-CBF has a higher violation rate at the initial training stage. With the estimation accuracy of GP model increasing, the violation rate gradually decreases.
It is worth mentioning that in  \cite{cheng2019end}, the estimation error bound in GPR-based uncertainty estimation is determined by a constant $k_\delta$ selected purely according to the desired confidence level, $1-\delta$. The computation of the error bound in this way is incorrect, and could potentially give an underestimation of the true error bound, especially when data is limited. The underestimated error bounds lead to high safety violation rates at the initial learning stage. A more rigorous approach for error bound determination is given in \cite{wallach_uniformboundGP}.


\begin{table*}[htb]
\vspace{-6mm}
\begin{center}
\caption{Safety violation rate during training for the unicycle}\label{table:safety_violation_uni}
\vspace{-2mm}
\begin{tabular}{|c|c|c|c|c|} 
\hline
Training Episode & 1$\sim$50 & 51$\sim$100 &  101$\sim$150  & 151$\sim$200\\
\hline
{DOB-CBF} & 0.0$\%$ & 0.0$\%$ & 0.0$\%$ & 0.0$\%$ \\ 
{GP-CBF} & 12.0$\%$ & 6.0$\%$ & 2.0$\%$ & 0.0$\%$ \\ 
\hline
\end{tabular}
\end{center}
\vspace{-8mm}
\end{table*}

\subsection{2D Quadrotor}
\begin{figure}[h]
\vspace{-1mm}
    \centering
\includegraphics[scale=0.4]{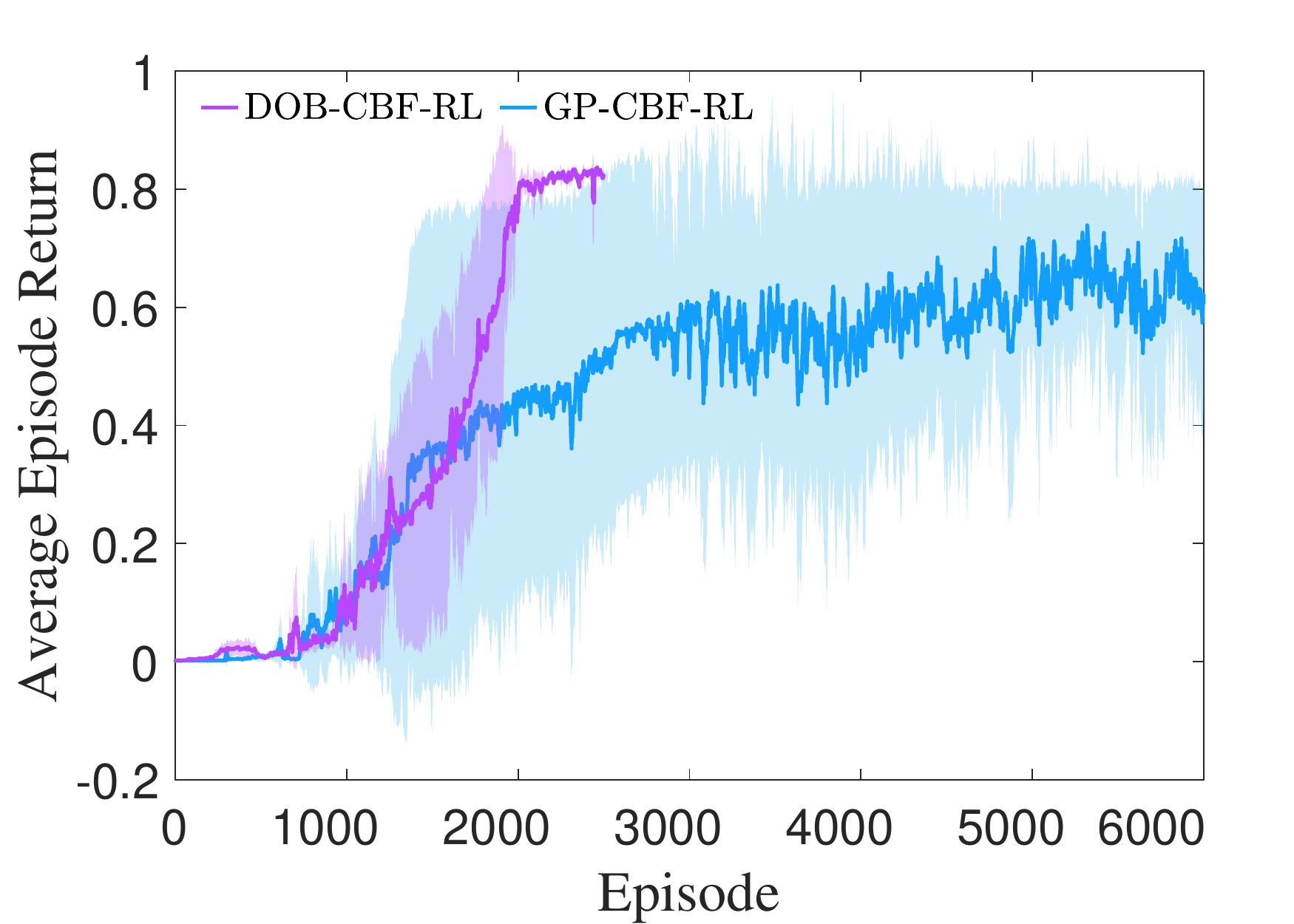}
\includegraphics[scale=0.4]{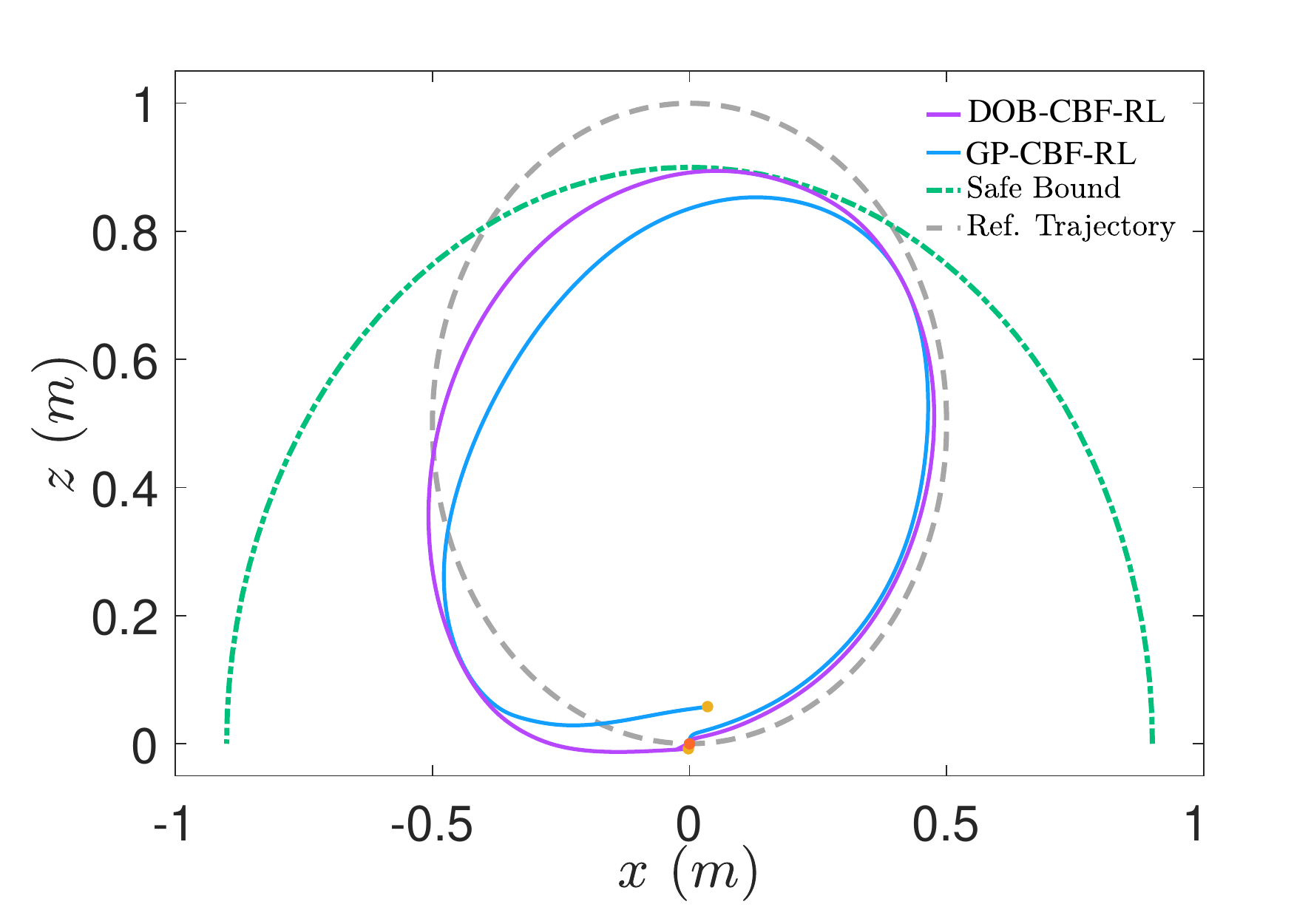}
\vspace{-3mm}
\caption{\black{(Left) Quadrotor training curves for DOB-CBF-RL and GP-CBF-RL. The solid lines and shaded areas denote the mean and standard deviation over five trials. The cumulative reward is normalized. (Right) Trajectory tracking performance for DOB-CBF-RL policy trained in 2500 episodes and GP-CBF-RL policy trained in 5000 episodes.}}\label{fig:quad_train_traj}
\vspace{-0mm}
\end{figure}
The state of the quadrotor is $x = [p_x, v_x, p_z, v_z, \theta,\dot{\theta}]^T$, where $[p_x, p_z]$ and $[v_x,v_z]$ are the position and velocity of the quadrotor in the $xz$-plane, respectively, and $[\theta, \dot{\theta}]$ are the pitch angle, i.e., the angle between $x$ direction of the quadrotor body frame and the $x$ direction of the inertia frame, and its angular velocity, respectively.
\black{The dynamics are given as follows:
\begin{equation}{\label{eq:quad_dyn}}
    \begin{aligned}
\dot{v_x} &=-\sin \theta\left(u_1+u_2+d_{u_1}+d_{u_2}\right) / m + d_{x},\\
\dot{v_z} &=\cos \theta\left(u_1+u_2+d_{u_1}+d_{u_2}\right) / m-g+ d_{z}, \\
\ddot{\theta} &=\left(u_2-u_1-d_{u_1}+d_{u_2}\right) L / I_{y y},
\end{aligned}
\end{equation}
where $g$ is the gravitational acceleration, $m$ denotes the total mass of the quadrotor, $L$ is the effective moment arm, $I_{yy}$ is the moment of inertia around $y$-axis, $d_{u_1}$ and $d_{u_2}$ are uncertainties to mimic the rotational friction of motors, and $d_{x}$ and $d_{z}$ denote the air resistance along each axis}.  
To be realistic, we impose the constrained control input $u_i\in[0,u_{\max} ]$ for $i=1,2$, where $u_{\max} = 2 $ N is the maximum thrust force generated by each rotor. The objective is to control the quadrotor to track a reference trajectory (denoted by the gray line in \cref{fig:quad_train_traj}) while staying within  a circle boundary with a radius $r_\textup{bnd} =0.85\text{m}$.   For DOB-CBF design,
we first chose a function $h(x) = \frac{1}{2}(r_\textup{bnd}^2-(p^2_x + p^2_y))$ as a candidate high-order CBF function for the nominal (i.e., uncertainty-free) system in the absence of control limits. The details about the dynamics, RL training and simulation settings can be found in \cite{cheng2022safe}. 
Figure \ref{fig:quad_train_traj} (Left) shows that the DOB-CBF-RL method can significantly improve the training efficiency, allowing the SAC policy to converge in less than two-thousand episodes. In any case, the GP-CBF-RL method failed to find an equally good policy in 6000 episodes in most trials. One can see from Figure \ref{fig:quad_train_traj} (Right) that DOB-CBF-RL enables the agent to generate a more aggressive trajectory. Knowing the accurate disturbance estimation, the policy trained with DOB-CBF-RL pushes the agent to finish the task as perfectly as possible, while still enforces the safety of the quadrotor.

\begin{figure}[h]
\vspace{-3mm}
    \centering
    \includegraphics[scale=0.37]{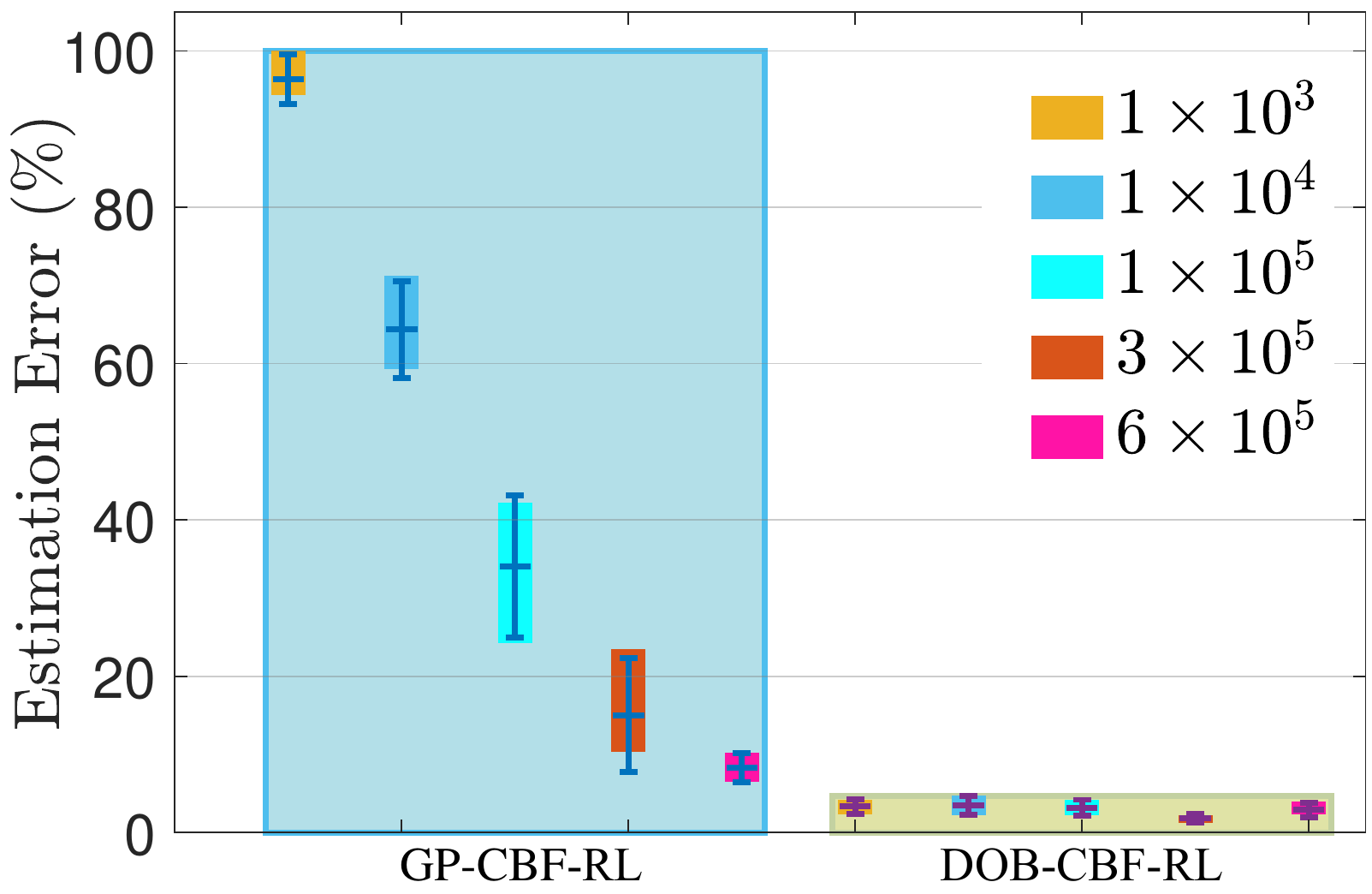}
    \hspace{3mm}\includegraphics[scale=0.36]{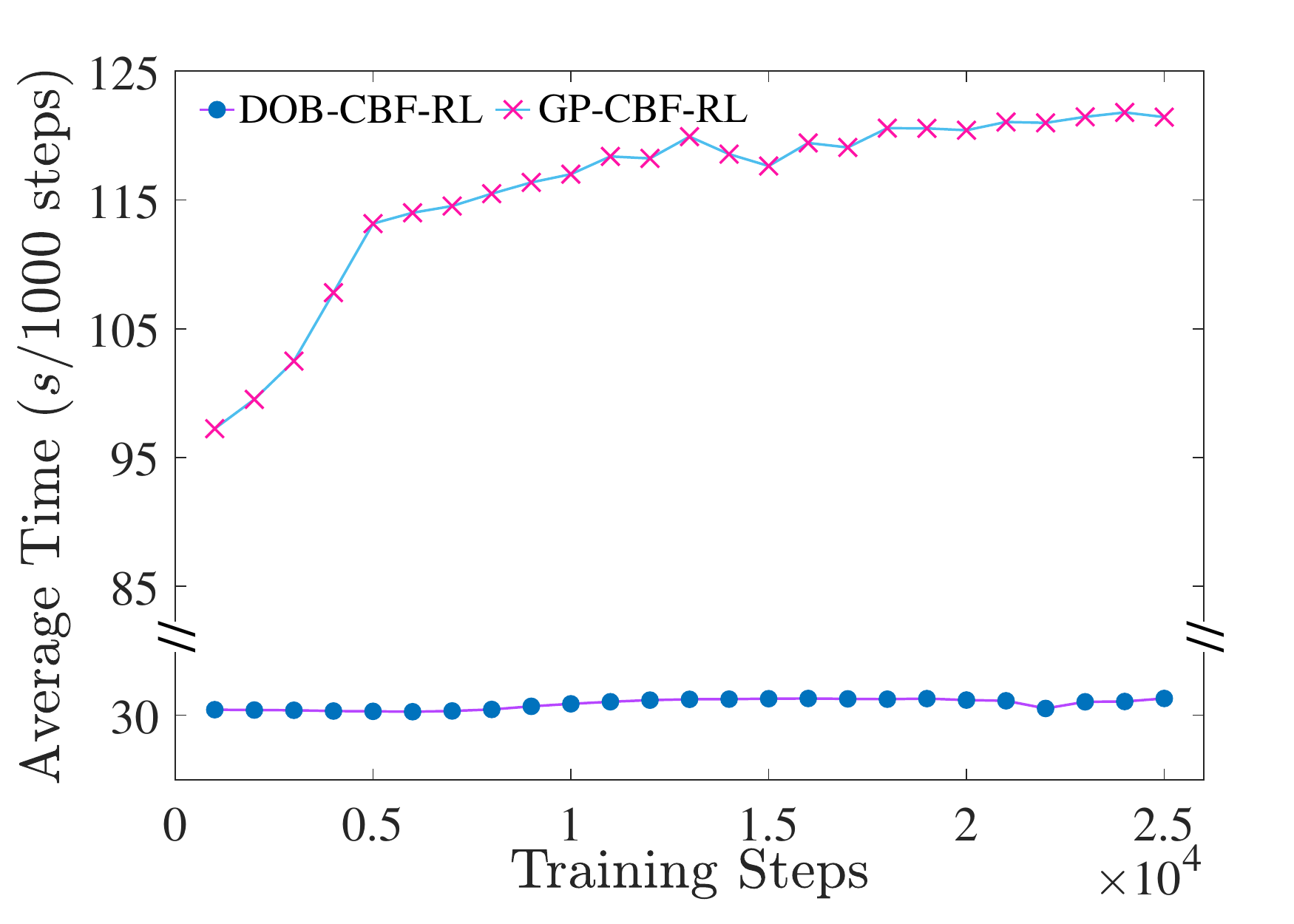}
\caption{\black{(Left) Disturbance estimation percentage error yielded by DOB and GP models at different steps during training. 
Five trials were performed and the mean with the standard deviation is shown as an error bar at each test step. The color box attached to each error bar denotes the worst and best estimation error at each shown step. For better illustration, we categorize all cases where the estimation percentage error is greater than 100\% as instances of 100$\%$ percentage error cases. (Right) Average computation time per 1000 steps. Solid lines with markers plot the average computation time per 1000 steps at $n$th training step}.
}\label{fig:quad_error_time}
\vspace{-2mm}
\end{figure}

Figure \ref{fig:quad_error_time} (Left) shows the disturbance estimation result at different training steps. DOB shows a relatively stable and decent estimation performance starting from the beginning and consistently yields an estimation error that is smaller than 5$\%$, while the GP model gradually decreases the error and yields larger estimation error even at $6\times 10^5$ steps. 
It is well known that GP model training involves computing a $N\times N$ covariance matrix $\Sigma$, where $N$ is the number of data points, which is computationally expensive when $N$ is large. Figure \ref{fig:quad_error_time} (Right) shows the average computation time per one thousand training steps, from which the computation time of GP-CBF-RL is at least about three times longer than the computation time of DOB-CBF-RL. To better validate the safe exploration feature, we compute the safety violation rates for every 500 episodes during training and summarize the results in Table \ref{table:safety_violation}. The {``w/ pre-training''} means we first trained a vanilla policy using nominal dynamics and used the pre-trained policy as the starting point for GP-CBF-RL. We can see from Table  \ref{table:safety_violation}, that DOB-CBF-RL shows an overwhelming advantage over the GP-CBF-RL method. Without pre-training, GP-CBF-RL shows significant safety guarantee performance at the initial training stage. In "w/ pre-training" case, GP-CBF still doesn't demonstrate evenly matched performance as DOB-CBF while its violation rates have been significantly lowered by introducing a pretrained policy. 
\black{Theoretically, DOB-CBF-RL is supposed to guarantee zero safety violations by leveraging a DOB-CBF function defined in \cref{defn:dob-cbf}. However, verifying whether a given function is a DOB-CBF, especially in the presence of control limits, is still a challenging problem. In other words, the intuitively selected function $h$ may not be a DOB-CBF in the presence of the uncertainties and control limits. As a result, the rigorous safety guarantee provided by our DOB-CBF-RL framework is lost. However, compared to GP-CBF-RL, our DOB-CBF framework still achieves much lower constraint violation rate throughout the learning phase.}


\begin{table*}[htb]
\vspace{-4mm}
\begin{center}
\caption{Safety violation rate during training for 2D quadrotor}\label{table:safety_violation}
\vspace{-2mm}
\begin{tabular}{|c|c|c|c|c|} 
\hline
Training Episode & 1$\sim$500& 501$\sim$1000&  1001$\sim$1500& 1501$\sim$2000\\
\hline
{DOB-CBF} & 15.8$\%$ & 2.6$\%$ & 0.2$\%$ & 0.0$\%$ \\ 
{GP-CBF} & 91.4$\%$ & 79.6$\%$ & 59.8$\%$ & 40.4$\%$ \\ 
{GP-CBF(w/ pre-training)} & 30.8$\%$ & 22.8$\%$ & 20.0$\%$ & 7.8$\%$  \\
\hline
\end{tabular}
\end{center}
\vspace{-10mm}
\end{table*}

\section{Conclusion and Future Work}
This paper presents a safe and efficient reinforcement learning (RL) scheme based on disturbance observers (DOBs) and control barrier functions (CBFs). Our approach leverages a DOB that can accurately estimate the pointwise value of the uncertainty,  and a quadratic programming (QP) module with a robust CBF condition, to generate safe actions by minimally modifying the (potentially unsafe) actions generated by the RL policy. Unlike existing safe RL approaches based on CBFs, which often rely on model learning of the uncertain dynamics, our approach completely removes the need for model learning and facilitates more sample- and computationally-efficient policy training. The efficacy of our proposed scheme is validated in simulated environments, in comparison with an existing CBF-based safe RL approach.

Our future work includes experimental validation of the proposed DOB-CBF-RL framework on a real robot, and extension of the framework to model-based RL settings.  
\section{Acknowledgments}
This work is supported in part by the AFOSR through Grant FA9550-21-1-0411, in part by the NASA through the ULI Grant 80NSSC17M0051, in part by NSF through the RI Grant 2133656 and the CMMI Grant 2135925, and in part by the HUJI/UIUC Joint Research and Innovation Seed Grant.
\bibliography{yikun,refs-pan}

\begin{thebibliography}{23}
\providecommand{\natexlab}[1]{#1}
\providecommand{\url}[1]{\texttt{#1}}
\expandafter\ifx\csname urlstyle\endcsname\relax
  \providecommand{\doi}[1]{doi: #1}\else
  \providecommand{\doi}{doi: \begingroup \urlstyle{rm}\Url}\fi

\bibitem[Ackerman et~al.(2019)Ackerman, Puig-Navarro, Hovakimyan, Cotting,
  Duke, Carrera, McCaskey, Esposito, Peterson, and
  Tellefsen]{ackerman2019Learjet}
Kasey Ackerman, Javier Puig-Navarro, Naira Hovakimyan, M.~Christopher Cotting,
  Dustin~J. Duke, Miguel~J. Carrera, Nathan~C. McCaskey, Dario Esposito,
  Jessica~M. Peterson, and Jonathan~R. Tellefsen.
\newblock Recovery of desired flying characteristics with an $\mathcal{L}_1$
  adaptive control law: Flight test results on {C}alspan's {VSS} {Learjet}.
\newblock In \emph{AIAA SciTech 2019 Forum}, San Diego, California, January
  2019.
\newblock {AIAA 2019-1084}.

\bibitem[Ackerman et~al.(2017)Ackerman, Xargay, Choe, Hovakimyan, Cotting,
  Jeffrey, Blackstun, Fulkerson, Lau, and Stephens]{ackerman2017evaluation}
Kasey~A. Ackerman, Enric Xargay, Ronald Choe, Naira Hovakimyan, M.~Christopher
  Cotting, Robert~B. Jeffrey, Margaret~P. Blackstun, T.~Paul Fulkerson,
  Timothy~R. Lau, and Shawn~S. Stephens.
\newblock Evaluation of an $\mathcal{L}_1$ adaptive flight control law on
  {C}alspan’s variable-stability {L}earjet.
\newblock \emph{AIAA Journal of Guidance, Control, and Dynamics}, 40\penalty0
  (4):\penalty0 1051--1060, 2017.

\bibitem[Alshiekh et~al.(2018)Alshiekh, Bloem, Ehlers, K{\"o}nighofer, Niekum,
  and Topcu]{alshiekh2018safe}
Mohammed Alshiekh, Roderick Bloem, R{\"u}diger Ehlers, Bettina K{\"o}nighofer,
  Scott Niekum, and Ufuk Topcu.
\newblock Safe reinforcement learning via shielding.
\newblock In \emph{Proceedings of the AAAI Conference on Artificial
  Intelligence}, volume~32, pages 2669--2678, 2018.

\bibitem[Ames et~al.(2016)Ames, Xu, Grizzle, and Tabuada]{ames2016cbf-tac}
Aaron~D Ames, Xiangru Xu, Jessy~W Grizzle, and Paulo Tabuada.
\newblock Control barrier function based quadratic programs for safety critical
  systems.
\newblock \emph{IEEE Trans. Automat. Contr.}, 62\penalty0 (8):\penalty0
  3861--3876, 2016.

\bibitem[Brunke et~al.(2022)Brunke, Greeff, Hall, Yuan, Zhou, Panerati, and
  Schoellig]{brunke2022safe}
Lukas Brunke, Melissa Greeff, Adam~W Hall, Zhaocong Yuan, Siqi Zhou, Jacopo
  Panerati, and Angela~P Schoellig.
\newblock Safe learning in robotics: {From} learning-based control to safe
  reinforcement learning.
\newblock \emph{Annual Review of Control, Robotics, and Autonomous Systems},
  5:\penalty0 411--444, 2022.

\bibitem[Chen et~al.(2015)Chen, Yang, Guo, and Li]{chen2015dobc}
Wen-Hua Chen, Jun Yang, Lei Guo, and Shihua Li.
\newblock Disturbance-observer-based control and related methods—-{An}
  overview.
\newblock \emph{IEEE Transactions on Industrial Electronics}, 63\penalty0
  (2):\penalty0 1083--1095, 2015.

\bibitem[Cheng et~al.(2019)Cheng, Orosz, Murray, and Burdick]{cheng2019end}
Richard Cheng, G{\'a}bor Orosz, Richard~M Murray, and Joel~W Burdick.
\newblock End-to-end safe reinforcement learning through barrier functions for
  safety-critical continuous control tasks.
\newblock In \emph{Proceedings of the AAAI Conference on Artificial
  Intelligence}, volume~33, pages 3387--3395, 2019.

\bibitem[Cheng et~al.(2022{\natexlab{a}})Cheng, Zhao, and
  Hovakimyan]{cheng2022safe}
Yikun Cheng, Pan Zhao, and Naira Hovakimyan.
\newblock Safe and efficient reinforcement learning using
  disturbance-observer-based control barrier functions.
\newblock \emph{arXiv preprint arXiv:2211.17250}, 2022{\natexlab{a}}.

\bibitem[Cheng et~al.(2022{\natexlab{b}})Cheng, Zhao, Wang, Block, and
  Hovakimyan]{cheng2022improvingRL-ral}
Yikun Cheng, Pan Zhao, Fanxin Wang, Jerome~Daniel Block, and Naira Hovakimyan.
\newblock Improving the robustness of reinforcement learning policies with
  $\mathcal{L}_{1}$ adaptive control.
\newblock \emph{IEEE Robotics and Automation Letters}, 7\penalty0 (3):\penalty0
  6574--6581, 2022{\natexlab{b}}.

\bibitem[Da{\c{s}} and Murray(2022)]{dacs2022robust-dob-cbf}
Ersin Da{\c{s}} and Richard~M Murray.
\newblock Robust safe control synthesis with disturbance observer-based control
  barrier functions.
\newblock In \emph{61st IEEE Conference on Decision and Control (CDC)}, pages
  5566--5573, 2022.

\bibitem[Emam et~al.(2021)Emam, Glotfelter, Kira, and Egerstedt]{SafembRL2022}
Yousef Emam, Paul Glotfelter, Zsolt Kira, and Magnus Egerstedt.
\newblock Safe model-based reinforcement learning using robust control barrier
  functions.
\newblock \emph{arXiv preprint arXiv:2110.05415}, 2021.

\bibitem[Fisac et~al.(2018)Fisac, Akametalu, Zeilinger, Kaynama, Gillula, and
  Tomlin]{fisac2018general-safety}
Jaime~F Fisac, Anayo~K Akametalu, Melanie~N Zeilinger, Shahab Kaynama, Jeremy
  Gillula, and Claire~J Tomlin.
\newblock A general safety framework for learning-based control in uncertain
  robotic systems.
\newblock \emph{IEEE Transactions on Automatic Control}, 64\penalty0
  (7):\penalty0 2737--2752, 2018.

\bibitem[Gahlawat et~al.(2020)Gahlawat, Zhao, Patterson, Hovakimyan, and
  Theodorou]{gahlawat2020l1gp}
Aditya Gahlawat, Pan Zhao, Andrew Patterson, Naira Hovakimyan, and Evangelos~A
  Theodorou.
\newblock {$\mathcal{L}_1$-$\mathcal{GP}$}: $\mathcal{L}_1$ adaptive control
  with {B}ayesian learning.
\newblock In \emph{Conference on Learning for Dynamics and Control}, volume
  120, pages 1--12, 2020.

\bibitem[Garc{\i}a and Fern{\'a}ndez(2015)]{garcia2015comprehensive}
Javier Garc{\i}a and Fernando Fern{\'a}ndez.
\newblock A comprehensive survey on safe reinforcement learning.
\newblock \emph{Journal of Machine Learning Research}, 16\penalty0
  (1):\penalty0 1437--1480, 2015.

\bibitem[Haarnoja et~al.(2018)Haarnoja, Zhou, Abbeel, and Levine]{Tuomassac}
Tuomas Haarnoja, Aurick Zhou, Pieter Abbeel, and Sergey Levine.
\newblock Soft actor-critic: Off-policy maximum entropy deep reinforcement
  learning with a stochastic actor.
\newblock In \emph{Proceedings of the 35th International Conference on Machine
  Learning}, volume~80, pages 1861--1870, 2018.

\bibitem[Hovakimyan and Cao(2010)]{naira2010l1book-nh}
Naira Hovakimyan and Chengyu Cao.
\newblock \emph{$\mathcal{L}_1$ Adaptive Control Theory: Guaranteed Robustness
  with Fast Adaptation}.
\newblock Society for Industrial and Applied Mathematics, Philadelphia, PA,
  2010.

\bibitem[Lederer et~al.(2019)Lederer, Umlauft, and
  Hirche]{wallach_uniformboundGP}
Armin Lederer, Jonas Umlauft, and Sandra Hirche.
\newblock Uniform error bounds for {Gaussian} process regression with
  application to safe control.
\newblock In \emph{Advances in Neural Information Processing Systems},
  volume~32, pages 657--667, 2019.

\bibitem[Nguyen and Sreenath(2016)]{nguyen2016optimalACC}
Quan Nguyen and Koushil Sreenath.
\newblock Optimal robust control for constrained nonlinear hybrid systems with
  application to bipedal locomotion.
\newblock In \emph{American Control Conference}, pages 4807--4813, 2016.

\bibitem[Ohnishi et~al.(2019)Ohnishi, Wang, Notomista, and
  Egerstedt]{ohnishi2019barrier}
Motoya Ohnishi, Li~Wang, Gennaro Notomista, and Magnus Egerstedt.
\newblock Barrier-certified adaptive reinforcement learning with applications
  to brushbot navigation.
\newblock \emph{IEEE Transactions on Robotics}, 35\penalty0 (5):\penalty0
  1186--1205, 2019.

\bibitem[Wabersich et~al.(2021)Wabersich, Hewing, Carron, and
  Zeilinger]{wabersich2021probabilistic-safety}
Kim~Peter Wabersich, Lukas Hewing, Andrea Carron, and Melanie~N Zeilinger.
\newblock Probabilistic model predictive safety certification for
  learning-based control.
\newblock \emph{IEEE Transactions on Automatic Control}, 67\penalty0
  (1):\penalty0 176--188, 2021.

\bibitem[Wang et~al.(2017)Wang, Yang, Sun, and Deng]{wang2017adaptiveMPC}
Xiaofeng Wang, Lixing Yang, Yu~Sun, and Kun Deng.
\newblock Adaptive model predictive control of nonlinear systems with
  state-dependent uncertainties.
\newblock \emph{Int. J. Robust Nonlinear Control}, 27\penalty0 (17):\penalty0
  4138--4153, 2017.

\bibitem[Xiao and Belta(2021)]{xiao2021high-order-cbf}
Wei Xiao and Calin Belta.
\newblock High order control barrier functions.
\newblock \emph{IEEE Transactions on Automatic Control}, 67\penalty0
  (7):\penalty0 3655--3662, 2021.

\bibitem[Zhao et~al.(2020)Zhao, Mao, Tao, Hovakimyan, and Wang]{zhao2020aR-cbf}
Pan Zhao, Yanbing Mao, Chuyuan Tao, Naira Hovakimyan, and Xiaofeng Wang.
\newblock Adaptive robust quadratic programs using control {Lyapunov} and
  barrier functions.
\newblock In \emph{59th IEEE Conference on Decision and Control}, pages
  3353--3358, 2020.

\end{thebibliography}

\end{document}